\documentclass{article}

\usepackage{arxiv}

\usepackage[utf8]{inputenc} 
\usepackage[T1]{fontenc}    
\usepackage{hyperref}       
\usepackage{url}            
\usepackage{booktabs}       
\usepackage{amsfonts}       
\usepackage{amsmath, amssymb}
\usepackage{nicefrac}       
\usepackage{microtype}      
\usepackage{lipsum}		
\usepackage{graphicx}
\usepackage[numbers]{natbib}
\usepackage{doi}
\usepackage{enumitem}
\usepackage{bm}
\usepackage{multirow}

\usepackage[table]{xcolor}

\definecolor{revs}{rgb}{0, 0, 0}

\title{Predicting Mechanically Driven Full-Field Quantities of Interest with Deep Learning-Based Metamodels}

\author{{Saeed Mohammadzadeh}\\
	Division of Systems Engineering\\
	Boston University\\
	Boston, MA 02215 \\
	\texttt{saeedmhz@bu.edu} \\
	\And
	{Emma Lejeune} \\
	Department of Mechanical Engineering\\
	Boston University\\
	Boston, MA 02215\\
	\texttt{elejeune@bu.edu} \\
}

\date{}

\renewcommand{\shorttitle}{\textit{arXiv}}

\hypersetup{
pdftitle={Predicting Mechanically Driven Full-Field Quantities of Interest with Deep Learning-Based Metamodels},
pdfsubject={q-bio.NC, q-bio.QM},
pdfauthor={Saeed Mohammadzadeh, Emma Lejeune},
pdfkeywords={First keyword, Second keyword, More},
}

\begin{document}
\maketitle

    \begin{abstract}
    	Using simulation to predict the mechanical behavior of heterogeneous materials has applications ranging from topology optimization to multi-scale structural analysis. However, full-fidelity simulation techniques such as Finite Element Analysis, while effective, can be prohibitively computationally expensive when they are used to explore the massive input parameter space of heterogeneous materials. Therefore, there has been significant recent interest in machine learning-based models that, once trained, can predict mechanical behavior at a fraction of the computational cost compared to full fidelity simulations. Over the past several years, research in this area has been focused mainly on predicting single Quantities of Interest (QoIs). However, there has recently been an increased interest in a more challenging problem: predicting full-field QoI (e.g., displacement/strain fields, damage fields) for mechanical problems. Due to the added complexity of full-field information, network architectures that perform well on single QoI problems may perform relatively poorly in the full-field QoI problem setting. This problem is also challenging because, even outside the Mechanics research community, deep learning approaches to image-to-image mapping and full-field image analysis remain poorly understood. The work presented in this paper is twofold. First, we made a significant extension to the Mechanical MNIST dataset designed to enable the investigation of full field QoI prediction. Specifically, we added Finite Element simulation results of quasi-static brittle fracture in a heterogeneous material captured with the phase-field method. This problem was chosen as a broadly relevant "challenge problem" for full-field QoI prediction. Second, we investigated multiple Deep Neural Network architectures and subsequently established strong baseline performance for predicting full-field QoI. We found that a MultiRes-WNet architecture with straightforward data augmentation achieves $0.80\%$ and $0.34\%$ Mean Absolute Percentage Error on full-field displacement prediction for Equibiaxial Extension and Uniaxial Extension datasets in the Mechanical MNIST Fashion dataset, respectively. In addition, we found that our MultiRes-WNet architecture combined with a basic Convolutional Autoencoder achieves a mean $F_1$ score of $0.87$ on the newly added Mechanical MNIST Crack Path dataset. In addition to presenting the results in this paper, we have released our model implementation and the Mechanical MNIST Crack Path dataset under open-source licenses. We anticipate that future researchers will directly use our model architecture on related datasets and potentially design models that exceed the baseline performance for predicting full-field QoI established in this paper.
    \end{abstract}

    \section{Introduction}
    \label{sec:intro} 
    Understanding and predicting the behavior of heterogeneous materials is a critical focus of mechanics researchers \citep{review-hetero-mat-modeling}. Over the past several decades, theoretical and computational advances have enabled high-accuracy simulation of heterogeneous materials behavior with applications ranging from patient-specific soft tissue simulation \citep{kakaletsis2021right,sahli2020classifying}, to material microstructure design \citep{bessa-design-metamaterial,design-kirigami,zhang2020machine}, to large scale structural analysis \citep{structural-analysis}. Full-fidelity modeling techniques such as the Finite Element Method \citep{finite-element-book} are typically quite effective, but they come with a major shortcoming: they are extremely computationally expensive \citep{rausch2017computational}, especially for advanced material models implemented on highly resolved meshes \citep{real-time-FE}. High computational cost often limits both our capability for thorough state-space exploration and prediction \citep{state-space-exploration}. On the other hand, metamodels, also referred to as surrogate models, are ``models of models'' that can replicate a portion of the results of the original high fidelity models through a portion of the state space with some (ideally low) associated error \citep{metamodeling1}. When metamodels are constructed with supervised machine learning methods, they are trained on data that is initially obtained from the outcomes of high fidelity models (e.g., Finite Element simulation) \citep{sid-kumar,sid-kumar2,vlassis2021sobolev}. Once trained, metamodels can then be used to make predictions on unseen data \citep{liang2018deep,mao2020designing}. 
    Machine learning methods such as Gaussian Process Regression \citep{gaussian-process-regression} and Convolutional Neural Networks \citep{cnn-review} have been used in mechanics-based metamodeling \citep{metamodeling-Gaussian-regression,cnn-metamodeling-mechanics}. Critically, these metamodels can be constructed to be computationally cheap \cite{wang2019meta}, which is massively enabling for applications such as topology optimization \citep{topology-optim}, multi-scale modeling \citep{multiscale-metamodeling,soft-tissue-lejeune,teichert2020scale}, and uncertainty quantification \citep{uncertainty-quantification}. However, despite the surge of interest in metamodeling for solid mechanics problems \citep{metamodeling-in-solid-mechancis}, most work has focused on predicting single Quantities of Interest (QoI) from descriptions of a material domain \citep{single-qoi-mechine-learning,gu2018novo,lejeune2021geometric}. In this work, we will address the more challenging problem of designing metamodels to predict full field QoI such as full field displacement, and full field damage \citep{failure-machine-learning-2,failure-machine-learning-3}. Notably, this has been a recently growing area of interest in the solid mechanics research field \citep{full-field-qoi-mechine-learning,failure-machine-learning-1,yang2021deep}. And, designing effective methods for full field QoI prediction is compelling because these methods: (1) leverage data that is typically already being generated by running full fidelity simulations for simulating single QoI problems, and (2) will lead to much richer predictions than single QoIs.

    In this work, our goal is to increase the impact of machine learning based metamodels in solid mechanics by designing methods for full field QoI prediction. 
    Full field QoI prediction exhibit some similarities with image-to-image translation tasks from the computer vision research field \citep{image-to-image-gan}.
    For example, image segmentation \citep{image-segmentation} and optical flow analysis \citep{optical-flow} are both computer vision tasks that can require full field predictions. 
    Despite the proliferation of methods for full field image analysis, it is only recently that researchers have begun adapting these methods for problems in mechanics \cite{umesh-liver}. 
    And, unlike in the computer vision field \citep{computer-vision-review}, there is a severe lack of open source datasets that can be used to explore problems in solid mechanics \citep{mechanics-datasets1,mechanics-datasets2}. 
    Therefore, the focus of this work is twofold: (1) on advancing full field QoI prediction in solid mechanics, and (2) on addressing the problem of lack of benchmark data for machine learning approaches to solid mechanics problems \citep{cmvis} by curating and disseminating open source datasets for full field QoI prediction. In recent previous work, we introduced the Mechanical MNIST dataset collection \citep{lejeune2019mechanicalUE,lejeune2020mechanicalEE,lejeune2020mechanical,lejeune2020mechanical0,lejeune2021exploring}, a series of open-source datasets, that includes Finite Element simulation results of applying a fixed set of loading and boundary conditions (e.g., uniaxial extension and equibiaxial extension) to a two-dimensional heterogeneous material domain. 
    In this work, we extend the Mechanical MNIST dataset collection by adding ``Mechanical MNIST Crack Path,'' a dataset of Finite Element simulation results of quasi-static brittle fracture in a heterogeneous material domain captured via the phase-field method \citep{phase-field-review,ambati-hybrid}. This significant extension is specifically designed as an interesting and challenging full field prediction problem because: 1) Computational models of fracture are typically quite complex and computationally expensive, and thus metamodeling approaches are a promising way to gain insight; and 2) Crack path prediction brings unique challenges that computer scientists and the machine learning community do not inherently address such as the requirement for high performance prediction of a small region of the image (i.e., the crack) that is not present in the original image. With our open source Mechanical MNIST dataset collection, we tackle the problem of full-field QoI prediction by proposing a Convolution-based deep learning framework, the ``MultiRes-WNet.'' This work presents not only a metamodeling approach that other researchers can readily implement, but also a framework for others to build directly on this work.
    
    The remainder of the paper is organized as follows. In Section \ref{sec:methods}, we first briefly discuss the previously published Mechanical MNIST and Mechanical MNIST Fashion datasets, and then introduce our new dataset, Mechanical MNIST Crack Path. Notably, our new dataset has two versions, a ``lite'' version designed for accessibility \citep{mech-mnsit-crack-path-openbu}, and an ``extended'' version designed so that other researchers can build on our work in a less constrained manner \citep{mech-mnsit-crack-path-dryad}. In Section \ref{sec:methods}, we also present the network architecture of the MultiRes-WNet, along with additional details of our metamodel implementation. Then, in Section \ref{sec:results}, we present the results of training our proposed metamodeling framework for both the displacement prediction task on the Mechanical MNIST and Mechanical MNIST Fashion datasets and the crack path prediction task on the Mechanical MNIST Crack Path dataset. Finally, in Section \ref{sec:conclusion}, we present concluding remarks and summarize possibilities for future work.
    
    \section{Methods}
        \label{sec:methods}
        As stated in \ref{sec:intro}, there are two main contributions presented in this paper. In Section \ref{sub:data generation}, we describe our contribution to the ``Mechanical MNIST'' dataset collection: ``Mechanical MNIST Crack Path.'' Then, in Section \ref{sec:metamodeling}, we present the details of our proposed novel metamodeling approach. We briefly note that additional information on the phase-field model used for dataset generation is presented in Appendix \ref{apx:phase_field}, and additional details of our metamodel implementation are presented in Appendix \ref{apx:metamodel}. 
        \subsection{Datasets and Data Generation} \label{sub:data generation}
        To test the efficacy of our proposed metamodeling technique, we need datasets to perform computational experiments. 
        In Section \ref{subsubsec:mech_m}, we briefly describe the full field QoI problem present in the original Mechanical MNIST dataset and the recently added Mechanical MNIST Fashion dataset. Then, in Section \ref{subsubsec:mech_m_cp}, we introduce our challenging addition to this collection, the ``Mechanical MNIST Crack Path'' dataset. {\color{revs}Visualizations of the different input parameter spaces in the Mechanical MNIST dataset collection are available in Appendix \ref{apx:umap}.}
        
        \subsubsection{Mechanical MNIST and Mechanical MNIST Fashion}
        \label{subsubsec:mech_m}
        Inspired by the classic MNIST dataset used by the computer vision research community \citep{mnist}, the original Mechanical MNIST and the Mechanical MNIST Fashion datasets \citep{lejeune2019mechanicalUE,lejeune2020mechanicalEE,lejeune2020mechanical} are benchmark datasets for comparing different metamodeling strategies on mechanical data. Each dataset contains Finite Element simulation results where a fixed set of loading and boundary conditions (e.g., uniaxial extension and equibiaxial extension) are applied to a two-dimensional heterogeneous material domain.  The heterogeneous material domain (a $28 \times 28$ unit square) in both datasets corresponds directly to bitmaps in the original MNIST and the Fashion MNIST datasets \citep{fashion-mnist}. The domain is modeled as a compressible Neo-Hookean material, and bitmap pixel intensity determines the elastic modulus of the corresponding region within the material domain. Therefore, each material block consists of a soft background matrix with a significantly stiffer region within. Both datasets contain the total change in the strain energy, total $x$ and $y$ reaction force, and full-field $x$ and $y$ displacements sampled on a $28\times28$ unit grid to match the size of the input material domain. In previous studies, we have investigated metamodels that predict single QoI (e.g., change in strain energy). In this study, we will focus exclusively on the full-field QoI prediction problem illustrated in Fig. \ref{fig:dataset}a and \ref{fig:dataset}b.
        
        \subsubsection{Mechanical MNIST Crack Path} 
        \label{subsubsec:mech_m_cp}
        Motivated by recent interest in designing machine learning models to predict material failure \citep{failure-machine-learning-1,failure-machine-learning-2,failure-machine-learning-3,failure-machine-learning-4,crack-ml-random-generation}, we decided to add a dataset to the Mechanical MNIST collection that captures crack propagation in a heterogeneous material. In particular, we were interested in designing a dataset where the goal of full field QoI prediction would be crack path prediction because this is a substantially different challenge than the full-field displacement prediction problem present in the original Mechanical MNIST dataset.
        For our new dataset, ``Mechanical MNIST Crack Path,'' we model quasi-static brittle fracture in a two dimensional heterogeneous material with inclusions whose positions are dictated by the Fashion MNIST bitmap using the phase-field method. 
        
        As a brief background, the phase-field method is a state-of-the-art approach to fracture modeling \citep{phase-field-review,zhang2016variational,ambati-hybrid,phase-field-miehe2010}. The core idea of the phase-field method is to approximate the discontinuity caused by a fracture with a damage field variable $\phi$. In this work, we follow the unified phase field method introduced by Wu in 2017 \citep{phase-field-unified}. This approach contains a general form for the crack density function $\gamma(\phi,\nabla\phi)$ which approximates the crack by a regularized functional $\Gamma_d(\phi)$, and is written as follows:
        \begin{equation} \label{equ:crack-approx}
            \Gamma \approx \Gamma_d(\phi) := \int_\Omega\gamma(\phi,\nabla\phi) \, \mathrm{d} \textbf{x} = \int_\Omega \frac{1}{c_0}\left(\frac{1}{l_0}\alpha(\phi)+l_0|\nabla\phi|^2\right) \mathrm{d} \textbf{x} \quad \mathrm{with} \quad c_0 = 4\int_0^1\sqrt{\alpha(\phi)}\, \mathrm{d}\phi,
        \end{equation}
        where $\alpha(\phi)$ is the geometric crack function and $l_0$ is an internal length scale parameter\cite{phase-field-unified}. We refer to Appendix \ref{apx:phase_field} for further details regarding the phase field theory and Finite Element implementation. We note briefly that, despite the benefits of the phase field method, it is quite computationally expensive because accurately capturing the crack geometry typically requires a very fine mesh surrounding the zone where the crack path could potentially propagate.
        
\begin{figure}
        	\centering
        	\includegraphics[width=\textwidth,height=\textheight,keepaspectratio]{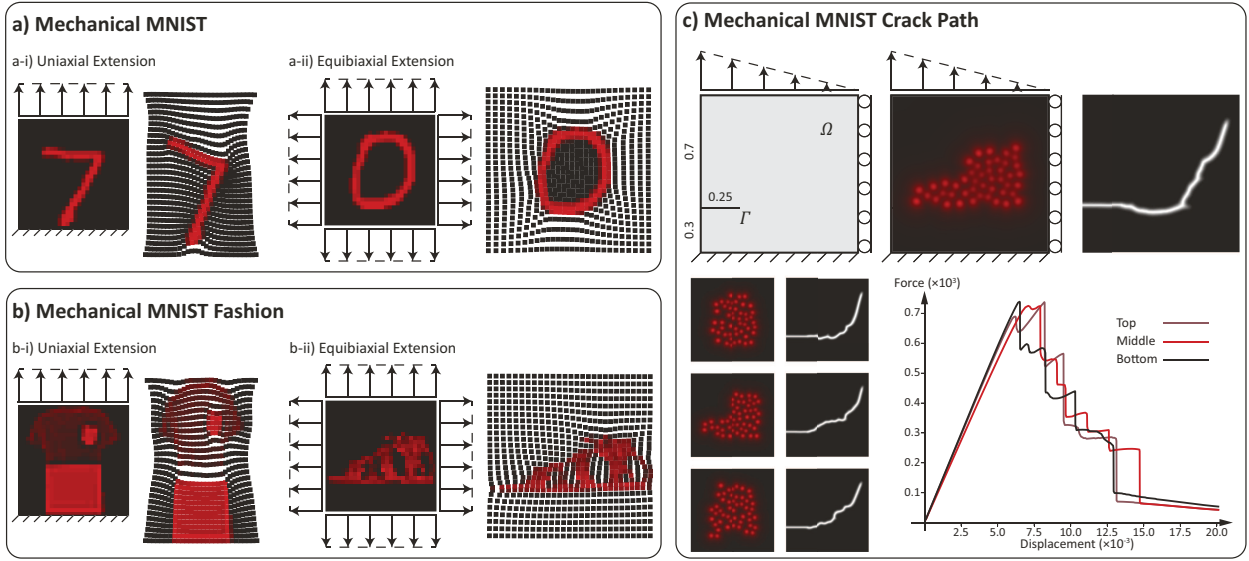}
        	\caption{Summary of the Mechanical MNIST datasets: a) and b) show the results of finite element simulations of uniaxial and equibiaxial extension for sample materials in the Mechanical MNIST and Mechanical MNIST Fashion datasets respectively; c) shows a schematic of the domain and boundary conditions defined for the Mechanical MNIST Crack Path dataset, as well as material distributions, damage fields and force vs. displacement curves for three sample cases. {\color{revs} A visualization of the three different input parameter domains in the Mechanical MNIST dataset collection is available in Appendix \ref{apx:umap}}}
        	\label{fig:dataset}
        \end{figure}

        In the Mechanical MNIST Crack Path dataset, illustrated in Fig. \ref{fig:dataset}c, we define the material domain as a square with a side length of $1$. There is an initial crack of fixed length ($0.25$) on the left edge of the domain. The bottom edge of the domain is fixed in $x$ (horizontal) and $y$ (vertical), the right edge of the domain is fixed in $x$ and free in $y$, and the left edge is free in both $x$ and $y$. The top edge is free in $x$, and in $y$ it is displaced such that, at each step, the displacement increases linearly from zero at the top right corner to the maximum displacement on the top left corner. Maximum displacement starts at $0.0$ and increases to $0.02$ by increments of $0.0001$ ($200$ simulation steps in total). The heterogeneous material distribution is obtained by adding very stiff circular inclusions to the domain using the Fashion MNIST bitmaps as the reference location for the center of the inclusions. Specifically, each center point location is generated randomly inside a square region (a fraction of the size of the domain) defined by the corresponding Fashion MNIST pixel when the pixel has an intensity value higher than $10$. In addition, a minimum center-to-center distance limit of $0.0525$ is applied while generating these center points for each sample. The values of Young’s Modulus $E$, Fracture Toughness $G_f$, and Failure Strength $f_t$ near each inclusion are increased with respect to the background domain by a variable rigidity ratio $r$. The background values for $E$, $G_f$, and $f_t$ are $210000$, $2.7$, and $2445.42$ respectively. The rigidity ratio throughout the domain depends on position with respect to all inclusion centers such that the closer a point is to the inclusion center the higher the rigidity ratio will be. The rigidity ratio is defined as:
        \begin{equation} \label{equ:rigidity-ratio}
            r(x) = 1+\left\{
                \begin{array}{cr}
                     3 & d < d_{min} \\
                     \frac{3d_{min}^2}{\beta d_{min}^2 + (1-\beta)d^2} & d \geq d_{min}
                \end{array}
                \right.
                \quad with \quad d = S_{\alpha}(||x-P_1||, \dots, ||x-P_n||)
        \end{equation}
        where $\alpha=-100$, $\beta=0.9$, $d_{min}=0.0075$, and $P_i$ is the center point of inclusion $i$. Function $S_\alpha$ is the smooth minimum function, and is defined as:
        \begin{equation} \label{equ:smooth-min}
            S_\alpha(x_1, \dots, x_n) = \frac{\sum_{i=1}^n x_i e^{\alpha x_i}}{\sum_{i=1}^n e^{\alpha x_i}} \quad where \quad S_\alpha \xrightarrow[]{} \min \; as \; \alpha \xrightarrow[]{} -\infty \, .
        \end{equation}
        We note that the full algorithm for constructing the heterogeneous material property distribution is included in the simulations scripts shared on GitHub, with access details given in Section \ref{sec:acknow}.
        \subsubsection{Note on the two versions of the Mechanical MNIST Crack Path dataset} 
        With this manuscript, we provide two separate versions of the Mechanical MNIST Crack Path dataset. We have designed a lite version of the dataset, available through the OpenBU Institutional Repository, such that others can download the dataset and directly use it to create metamodels without any additional modification. In addition, a full version of the dataset, available through the Dryad Repository, contains more information but will require preprocessing before it can be used to train metamodels. Notably, the full version of the dataset allows for more flexibility in metamodeling, and is intended to be used by researchers who are familiar with the underlying mechanical problem. The OpenBU version of the dataset contains:
        \begin{itemize}[noitemsep]
            \item An array corresponding to the rigidity ratio to capture heterogeneous material distribution reported over a uniform $64\times64$ grid. 
            \item The binary damage field at the final level of applied displacement reported over a uniform $256\times256$ grid. 
            \item The force-displacement curves for each simulation (which are not addressed in this manuscript). 
        \end{itemize}
        The larger version of the dataset contains: 
        \begin{itemize}[noitemsep]
            \item The locations of the center of each inclusion, where the script to subsequently extract rigidity ratio matrices with a desired resolution is available on GitHub. 
            \item The displacement and damage fields every ten simulation steps reported over a uniform $256\times256$ grid. 
            \item The full mesh resolution displacements and damage fields at both the last step of the simulation and at the early stage of damage initiation (when the reaction force is maximum). 
            \item The force-displacement curves for each simulation. 
        \end{itemize}
        \subsection{Metamodeling} 
        \label{sec:metamodeling} 
        In this Section, we briefly summarize the metamodel architecture that we developed for predicting full field QoI for mechanics based problems. In Section \ref{subsubsec:mrwn}, we introduce the MultiRes-WNet architecture, in Section \ref{subsubsec:autoencoder} we describe our autoencoder based approach to managing the high domain resolution required for crack path prediction, and in Section \ref{subsubsec:meta_alt} we briefly note alternative approaches that were (in our experience) less effective than our final metamodel design. {\color{revs}In Appendix \ref{apx:metamodel}, we provide a quantitative comparison between our proposed MultiRes-WNet and three representative alternatives.}
        
        \subsubsection{MultiRes-WNet Network Architecture}
        \label{subsubsec:mrwn}
\begin{figure}
        	\centering
        	\includegraphics[width=\textwidth,height=\textheight,keepaspectratio]{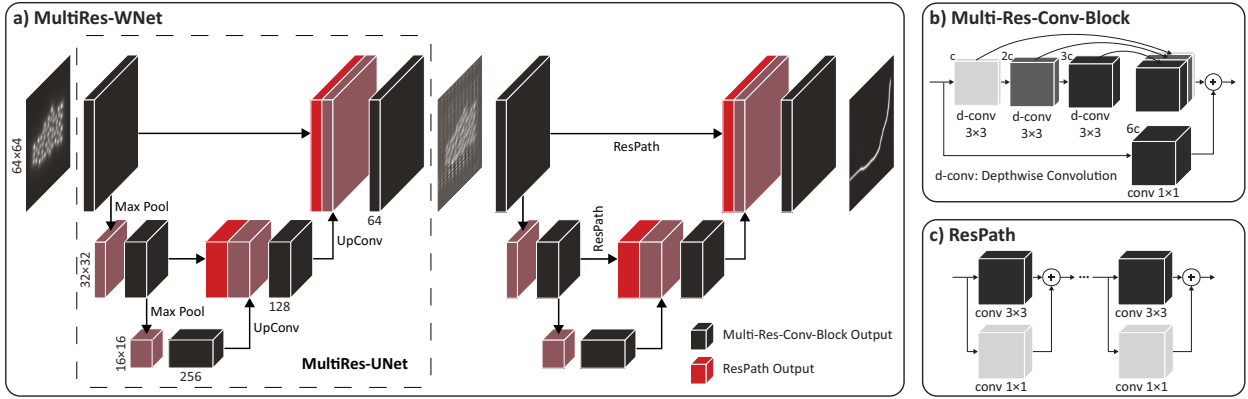}
        	\caption{We combined the U-Net with Multi-resolution Convolutional Blocks, ResPath, and Depthwise Separable Convolutions to form MultiRes-WNet, which consists of two identical MultiRes-UNet models put together in series. We schematically illustrate: a) the MultiRes-WNet Network Architecture; b) the Multi-resolution Convolutional Block that replaces regular convolutions in the U-Net to enable feature extraction at different zoom levels; and (c) the ResPath that applies a few convolutions on the input with low-level features from the encoder to match the output with rich features in the decoder section.}
        	\label{fig:multires-wnet}
        \end{figure}

        The U-Net \citep{unet} architecture is a type of Convolutional Neural Network (CNN) developed mainly for image segmentation tasks \citep{unet-seg-1,unet-seg-2,unet-seg-3}. The U-Net has been widely used in medical image analysis \citep{unet-med-seg-1,unet-med-seg-2,unet-med-seg-3}. Furthermore, in recent years, the U-Net architecture has been adopted in several areas of solid mechanics research for full-field QoI prediction, such as modeling the non-linear relation between a contact force and the displacement field in hyperelastic material domains \citep{umesh-liver}, and full-field stress and strain predictions in composite geometries \citep{unet-strain-stress}. In fluid mechanics, the U-Net architecture has also been used for velocity and pressure field prediction \citep{unet-fluid1}, optimal mesh generation \citep{unet-fluid2}, and correcting numerical error induced by a coarse-grid simulation of turbulent flows \citep{unet-fluid3}. The U-Net architecture consists of a downsampling (encoder) section to extract high-level features from the input image, and an upsampling (decoder) section to construct the desired output. In a typical implementation, the encoder takes the input array (image or otherwise), and applies a series of convolutions followed by ReLU activation functions, followed by max pool operations for each level of downsampling. The decoder section of the U-Net takes inputs at each level that are constructed by concatenating the encoder output of the same level with a transpose convolution of the lower level encoder output (referred to in short as an ``UpConv''). This basic ``U'' motif can be seen in Fig. \ref{fig:multires-wnet}a. 
        
        In this work, after our initial dissatisfaction with the efficacy of the standard U-Net, we propose a modified version of the U-Net architecture for predicting full field mechanics based QoI. The key modifications that we applied to the UNet architecture are summarized as follows: 1) we stack two UNets in series, 2) we use depthwise separable convolutions, 3) we replace double convolutions at each layer with a block of serial convolutions with internal residual paths, and 4) we add several convolutional filters to the residual path connecting encoder output to the decoder input. Our first type of major modification the the UNet, the ``MultiRes'' architecture, was initially proposed as a MultiRes-UNet \citep{multiresunet} with two key modifications to the standard U-Net architecture: 1) replacing the sequence of regular convolutions by a multi-resolution convolutional block (MultiRes-Conv-Block), and 2) passing the encoder output at each level through a convolutional residual path (ResPath) before concatenation with the upsampled tensor. These modifications are schematically illustrated in Fig. \ref{fig:multires-wnet}b and Fig. \ref{fig:multires-wnet}c. The addition of the MultiRes-Conv-Block enables feature extraction at different zoom levels, and the intuition behind the ResPath modification is that it avoids direct concatenation of low-level features in the encoder section with rich features in the decoder section of the network. In addition to the initial major modification, we replaced the regular convolutions in the MultiRes-Conv-Block with Depthwise Separable Convolutions \citep{depthwise-separable-convs}. The idea of these is to first apply separate convolutional filters on each channel, and then apply a point-wise filter that combines values of a pixel over all channels. Finally, inspired by \citep{wnet-kulis} and \citep{WNet-aerial-segmentation}, we stack two MultiRes-UNets with depthwise separable convolutions in series. We found through experiments that doubling the number of parameters with this approach is more effective than doubling the number of channels in a single UNet. Because of this addition, we refer to our network as a "MultiRes-WNet," illustrated in Fig. \ref{fig:multires-wnet}a. \\
        
        \subsubsection{An autoencoder based approach to handling the high output domain resolution requirements for accurate crack path prediction}
        \label{subsubsec:autoencoder}
\begin{figure}
        	\centering
        	\includegraphics[width=0.65\textwidth,height=\textheight,keepaspectratio]{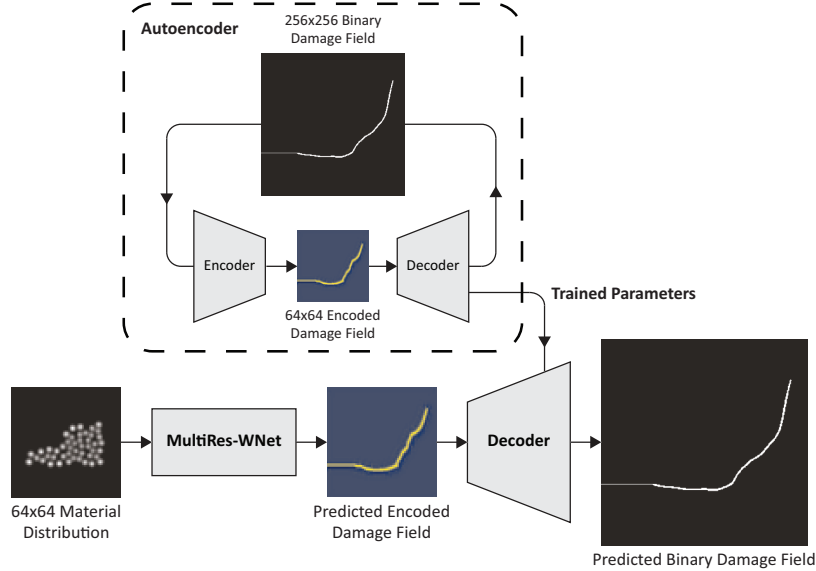}
        	\caption{A schematic illustration of the MultiRes-WNet combined with an autoencoder for crack path prediction. The autoencoder encodes the high-resolution damage field to a low-resolution array, and the MultiRes-WNet learns to map the low-resolution material distribution to low-resolution encoded damage field. The decoder of the trained autoencoder takes the predicted low-resolution encoded damage field and outputs the predicted high-resolution damage field.}
        	\label{fig:autoencoder}
        \end{figure}

        For the Mechanical MNIST and MNIST Fashion datasets, the full field QoI prediction challenge involves mapping the $28 \times 28$ material property input domain to the full-field displacement output on a $28 \times 28$ grid (see Section \ref{subsubsec:mech_m} for details). For this scenario, the MultiRes-WNet described in Section \ref{subsubsec:mrwn} requires no additional modification. However, for the Mechanical MNIST Crack Path dataset, a $256 \times 256$ array is required for the damage field output domain to sufficiently capture the crack path geometry. Also, based on the network architecture we use, the resolution of the material distribution input domain must match the resolution of the output domain.  
        In this scenario, we run into a potential limitation of the MultiRes-WNet architecture: an unacceptably high training time. Quadrupling the resolution of the input images from $64\times64$ to $256\times256$ results in an approximately ten to fourteen times greater training time when using consistent computational resources. {\color{revs} With the resources presently available to us (two NVIDIA TESLA V100 GPUs), training time for a $64 \times 64$ input domain is $\approx 1$ day, therefore an order of magnitude increase in training time would be highly detrimental.} In addition, because the output domain damage variable is equal to zero over most of the problem domain (everywhere except near the crack), using Mean Squared Error (MSE) loss will often result in poor prediction performance {\color{revs}where the network ultimately predicts a severely smeared, rather than binary, damage field.} To address these challenges, we combine our MultiRes-WNet architecture with a basic Convolutional Autoencoder \citep{autoencoder} to reduce the size of the damage field arrays for both faster and more accurate training of the MultiRes-WNet, illustrated in Fig. \ref{fig:autoencoder}. Specifically, we first train the convolutional autoencoder on a subset of the training data to effectively transform a ($256 \times 256$) damage field into a ($64 \times 64$) encoded damage field, and then reconstruct the true ($256 \times 256$) damage field with a negligible error, illustrated in Fig. \ref{fig:autoencoder}. Once the autoencoder is trained, we first encode the damage field for the entire training set, and then train the MultiRes-WNet to predict the encoded damage field. As illustrated in Fig. \ref{fig:autoencoder}, the MultiRes-WNet will map the material property distribution to the encoded damage field. Then, we reconstruct the predicted high-resolution damage field with the previously trained decoder section of the autoencoder. {\color{revs} This simple to implement modification provided us with a pragmatic way of dealing with the aforementioned drawbacks of training a MultiRes-WNet directly on the unencoded representation of the binary damage field. We anticipate that alternative approaches could be similarly effective and we make no claim that the use of an autoencoder is the optimal approach to this problem.}

    \subsubsection{Notes on metamodel design process and alternative approaches investigated}
    \label{subsubsec:meta_alt}
    As stated briefly in \ref{subsubsec:mrwn}, we used the UNet model in our first attempt at designing an effective metamodel for full-field QoI prediction problems. This initial exploration focused on displacement field prediction for the Mechanical MNIST datasets. Specifically, we tested the original UNet and several modified versions of the UNet: the Channel-Unet \citep{channel-unet}, the Attention U-Net \citep{attention-unet}, the parallel and coupled UNet architectures \citep{unet-fluid1}, the W-Net \citep{wnet-kulis}, and multiple combinations of these modifications. Although there is certainly more room for exploration, we found that the MultiRes-WNet was the best performing metamodel in our experiments. Once we had settled on the Multires-Wnet architecture as a promising method for predicting displacement fields, we tested the MultiRes-WNet on an inherently different task of crack path prediction using the Mechanical MNIST Crack Path dataset. Although it was necessary to use an autoencoder to handle the high-resolution data requirements outlined in \ref{subsubsec:autoencoder}, the MultiRes-WNet performed well on the crack path prediction task. However, despite our promising results, we believe that metamodeling techniques in this field are still far from perfect, and we expect future studies to focus on designing metamodels that exceed the current state of the art performance established in this paper for full-field QoI prediction on the Mechanical MNIST, Mechanical MNIST Fashion, and Mechanical MNIST Crack Path benchmark datasets. 
    
    \section{Results and Discussion}
    \label{sec:results}
    \subsection{Predicting Full Field Displacement for the Mechanical MNIST and Mechanical MNIST Fashion Datasets} 
    \label{subsec:res1}
    In order to evaluate the performance of our proposed MultiRes-WNet architecture, illustrated in Fig. \ref{fig:multires-wnet}, we trained the MultiRes-Wnet on all $60000$ samples provided in the Mechanical MNIST Uniaxial Extension, Mechanical MNIST Equibiaxial Extension, Mechanical MNIST Fashion Uniaxial Extension, and Mechanical MNIST Fashion Equibiaxial Extension datasets. 
    As stated in Section \ref{subsubsec:mech_m}, these datasets  map $28\times28$ material property distribution arrays to $28\times28$ $x$ and $y$ displacement fields. 
    For all metamodel training, we set the learning rate to $10^{-2}$ with a decay rate of $0.5$ at every $50$ epochs. We set the batch size to $32$, and the number of channels in the first layer to $64$, which corresponds to approximately $2$ million total metamodel parameters.
    In addition, we trained each model for $200$ epochs using Adam optimizer. Throughout this Section, we evaluate each trained metamodel on all $10000$ available test cases, and report the absolute error in displacement prediction for each test case averaged over the $28\times28$ grid. In addition, we divide the absolute error value by the mean nodal displacement averaged over all samples in each dataset to obtain a dimensionless Absolute Percentage Error ($APE$) metric for fair performance comparison between the different datasets that may have different levels of applied displacement. 
    
    \begin{figure}
    	\centering
    	\includegraphics[width=\textwidth,height=\textheight,keepaspectratio]{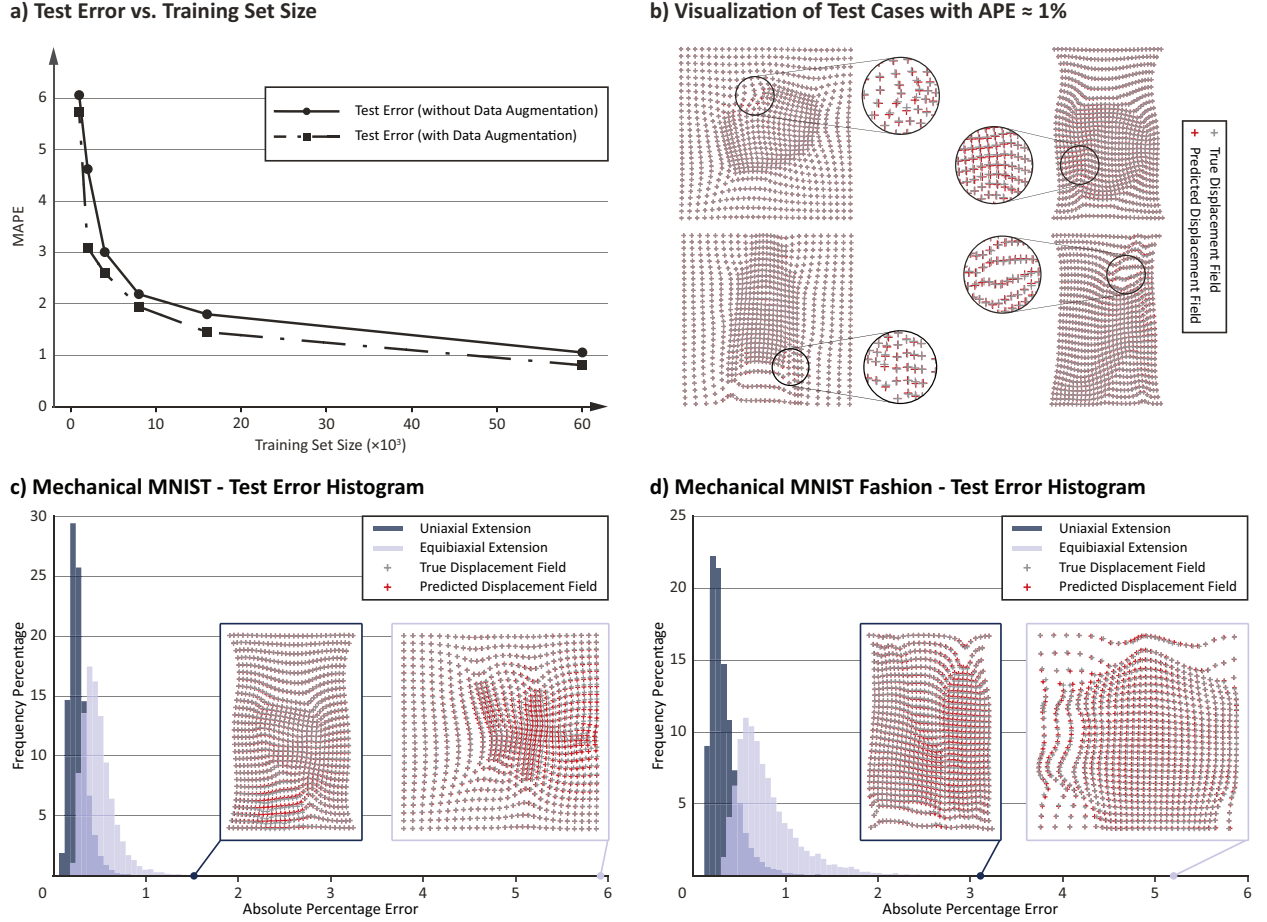}
    	\caption{Metamodel performance evaluation for displacement prediction: a) MAPE vs. training set size for models trained on subsets of the Mechanical MNIST Fashion Equibiaxial Extension Dataset (with and without data augmentation) and tested on the Mechanical MNIST Fashion test set; b) Sample visualizations of the predicted vs. true displacements for test cases with $APE \approx 1.0\%$; c) Histograms of the frequency of the predicted displacement field absolute percentage error for Mechanical MNIST Uniaxial Extension and Equibiaxial Extension test datasets; and d) Mechanical MNIST Fashion Uniaxial Extension and Equibiaxial Extension test datasets. In c) and d) the illustrations of the predicted vs. true displacement field are the worst performing example from each test dataset.}
    	\label{fig:displacement-pred-histogram}
    \end{figure}

    In Fig. \ref{fig:displacement-pred-histogram}a, we show the effect of both training set size and data augmentation on metamodel test performance for the Mechanical MNIST Fashion Equibiaxial Extension dataset. 
    We note briefly that for both Mechanical MNIST Fashion datasets, we performed both horizontal and vertical flips (separately and combined) to augment the size of the initial training set to $4\times60000$ training points. Because the original Mechanical MNIST datasets were ``easier'' (i.e., lower metamodel test error) we did not perform data augmentation for either of them. We note that the results presented in Fig. \ref{fig:displacement-pred-histogram}a show that data augmentation consistently improves the model performance, and that at least $10000$ training cases are required for our model to effectively learn the distribution of the test set.
    For the metamodels trained with all $60000$ training points (including data augmentation for the Mechanical MNIST Fashion datasets), metamodel performance is summarized in Table \ref{tab:disp-pred-results}. Briefly, we observed a Mean Absolute Percentage Error (MAPE) of $0.27\%$ and $0.34\%$ for Uniaxial Extension and $0.50\%$ and $0.80\%$ for Equibiaxial Extension in the Mechanical MNIST and Mechanical MNIST Fashion datasets, respectively. Qualitatively, as shown in Fig. \ref{fig:displacement-pred-histogram}b, for a single test case with $APE < 1\%$, the true and predicted displacement fields are visually indistinguishable. If $APE= 1\%$ is considered the cutoff for ``high quality'' predictions, our metamodels achieve $99.87\%$ and $97.50\%$ of high quality predictions on the Mechanical MNIST Uniaxial and Equibiaxial Extension datasets and $99.25\%$ and $78.84\%$ high quality predictions on the Mechanical MNIST Fashion Uniaxial and Equibiaxial Extension datasets respectively. The distributions of test error for all four datasets are shown in Fig. \ref{fig:displacement-pred-histogram}c and Fig. \ref{fig:displacement-pred-histogram}d. In addition to showing the distribution of test error, we added visualization of the \textit{worst} prediction (i.e., highest test error) from each of the four datasets. These results show that the MultiRes-WNet architecture is quite effective for predicting full field displacement on all four datasets, and is thus likely a good choice for related problems in mechanics. {\color{revs}In Appendix \ref{apx:comparison-disp-pred}, we compare three alternative Neural Network architectures with the MultiRes-WNet, and, as shown in Table \ref{tab:disp-pred-alternatives}, the MultiRes-WNet outperforms all of these alternatives. In addition, we analyze the performance of the MultiRes-WNet in strain field prediction in Appendix \ref{apx:strain-pred}. In short, the MultiRes-WNet is an effective approach for predicting strain fields in addition to displacement fields.}
    
    \begin{table}[htbp] \label{tab:disp-pred-results}
      \centering
      \caption{MultiRes-WNet performance predicting full field displacement on the Mechanical MNIST and Mechanical MNIST Fashion datasets. All values reported are test error.}
        \resizebox{0.75\textwidth}{!}{\begin{tabular}{cccccc}
        \toprule
              &       & \multicolumn{2}{c}{\textbf{Mechanical MNIST}} & \multicolumn{2}{c}{\textbf{Mechanical MNSIT Fashion}} \\
    \cmidrule{3-6}          &       & \textbf{Uniaxial Extension} & \textbf{Equibiaxial Extension} & \textbf{Uniaxial Extension} & \textbf{Equibiaxial Extension} \\
        \midrule
        \textbf{X Displacements} & \multirow{3}[2]{*}{\textbf{MAE}} & 0.0101 & 0.0128 & 0.0132 & 0.0229 \\
        \textbf{Y Displacements} &  & 0.0129 & 0.130 & 0.0149 & 0.0204 \\
        \multirow{2}[3]{*}{\textbf{Total Displacements}} &  & 0.0181 & 0.0205 & 0.0222 & 0.0344 \\
    \cmidrule{2-6}          & \textbf{MAPE} & 0.27\% & 0.50\% & 0.34\% & 0.80\% \\
        \bottomrule
        \end{tabular}}
    \end{table}%
    
    \begin{table}[htbp]
      \centering
      \caption{MultiRes-WNet performance predicting crack path on the Mechanical MNIST Crack Path. All values reported are for the test.}
        \resizebox{0.6\textwidth}{!}{\begin{tabular}{cccc}
        \toprule
              & \multicolumn{3}{c}{\textbf{Mechanical MNIST Crack Path}} \\
    \cmidrule{2-4}          & \textbf{Continuous Path} & \textbf{Discontinuous Path} & \textbf{All Samples} \\
        \midrule
        \textbf{Number of Samples} & 8861  & 1139  & 10000 \\
        \textbf{F1 Score} & 0.893 & 0.693 & 0.870 \\
        \textbf{Avg. Wrong Pixel Count (FP+FN)} & 147   & 411   & 177 \\
        \bottomrule
        \end{tabular}%
      \label{tab:crack-path}}%
    \end{table}%

    \subsection{Predicting Full Field Damage for the Mechanical MNIST Crack Path Dataset}
    \label{subsec:res2-crac}
    As stated in Section \ref{subsubsec:mech_m_cp}, we created the Mechanical MNIST Crack Path dataset as an additional challenge for our proposed MultiRes-WNet metamodel. In addition to the MultRes-WNet, we also used a convolutional autoencoder to reduce the resolution of the $256\times256$ array required to resolve the crack path captured by the phase field model damage field to a $64\times64$ array (see Section \ref{subsubsec:autoencoder} for details). Therefore, we trained our MultiRes-WNet to predict a $64\times64$ encoded array from a $64\times64$ material property distribution input array. Before training our MultiRes-WNet, we train the autoencoder on $10,000$ samples randomly selected from the training set of the Mechanical MNIST Crack Path dataset. {\color{revs}Note that in Appendix \ref{apx:comparison-1d-crack}, we evaluate an alternative method for crack prediction based on a 1D representation of the crack path and found that it was less effective than our proposed autoencoder approach.}
    
    Our trained autoencoder is able to regenerate true damage fields with an average accuracy of $99.995\%$ (i.e., $2.74$ incorrect pixels out of $65536$ per sample). With the trained autoencoder, we generated encoded damage fields for $60000$ samples in the training set and all $10000$ samples in the test set. Then, after we mapped the $64\times64$ material property distribution arrays to $64\times64$ encoded damage fields with our MultiRes-WNet, we used the trained decoder to generate the $256\times256$ predicted damage fields. To evaluate our model, we set a damage field threshold $(\phi > 0.5)$ and thus treat the predicted damage field as a binary image, in which pixel value $=1$ denotes the crack path (i.e., the representation of the crack path).
    
    We train our MultiRes-WNet on all $60000$ training samples provided in the Mechanical MNIST Crack Path dataset. We set the learning rate to $10^{-2}$ with a decay rate of $0.5$ at every $25$ epochs and trained the model for $100$ epochs in total using Adam optimizer. We set the batch size to be $32$ and the number of channels in the first layer to be $64$. We evaluated the trained metamodel on the $10000$ test cases provided in the dataset. To quantitatively evaluate the model performance, we used the S\o{}rensen–Dice index, referred to as the $F_1$ score and defined as:
    \begin{equation} \label{equ:dice-score}
        F_1 = \frac{2TP}{2TP + FP + FN},
    \end{equation}
    where $TP$, $FP$, and $FN$ are the total number of true positive, false positive and false negative pixels in a predicted sample. We note that the number of true negative pixels does not affect this score which is desirable because the number of true negative pixels is two orders of magnitude larger than $TP$, $FP$, and $FN$ pixels combined.
    
    \begin{figure}[ht]
    	\centering
    	\includegraphics[width=\textwidth,height=\textheight,keepaspectratio]{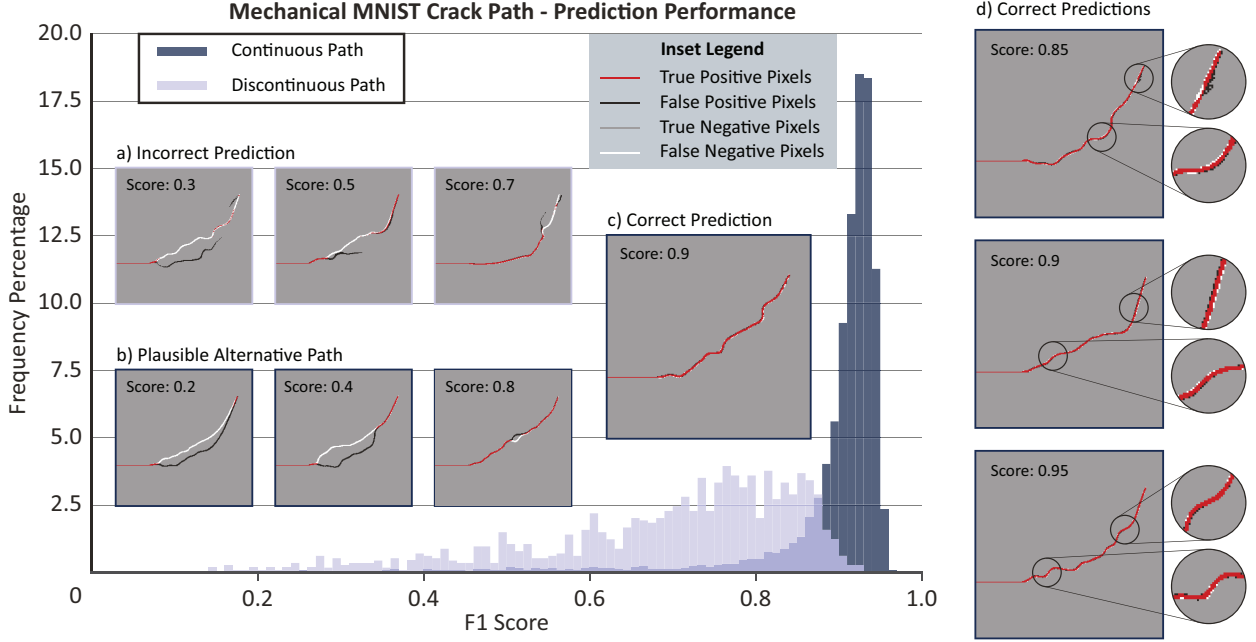}
    	\caption{\label{fig:crackres} Metamodel performance evaluation on the Mechanical MNIST Crack Path dataset. The histograms show the frequency of test set results with respect to $F_1$ score interval (see Eqn. \ref{equ:dice-score}). The predicted samples are grouped into continuous (dark blue) and discontinuous (light blue) paths. To contextualize these scores, we show: a) Three discontinuous paths with different scores (referred to as ``Incorrect Predictions''); b) Three continuous paths with different scores (referred to as ``Plausible Alternative Paths''); and c) A sample continuous path with a relatively high F1 score (referred to as a ``Correct Prediction''). To contextualize model performance, we consider a continuous paths with a F1 score higher than $0.85$ a ``Correct Prediction'' for this dataset. d) Three samples of correctly predicted crack paths with $F_1$ scores of $0.85$, $0.90$, and $0.95$.}
    \end{figure}
    
    As illustrated in Fig. \ref{fig:crackres}, we categorize the predicted crack path patterns into continuous and discontinuous paths. Notably, for the Mechanical MNIST Crack Path dataset, all predicted crack paths \textit{should} be continuous. In Fig. \ref{fig:crackres}, we show histograms of the $F_1$ score of the predicted damage field for the test data separated by category. To help qualitatively interpret these results, we consider damage field predictions as ``Correct'' if the predicted path is continuous, and if the corresponding $F_1$ score is above a threshold of $0.85$. We show an example of ``Correct'' behavior in Fig. \ref{fig:crackres}c. For continuous predictions with a low $F_1$ score (below $0.85$), we observed that the crack path tends to follow a  ``Plausible Alternative Path,'' where the crack propagates around inclusions in an alternative manner to the true behavior. We show three examples of this behavior in Fig. \ref{fig:crackres}b. {\color{revs} In Appendix \ref{apx:plausible-alt}, we discuss this qualitative categorization in more detail and show the sensitivity of our results to the chosen ``Plausible Alternative Path'' threshold.}  Notably, a possible explanation for the existence of plausible alternative paths is the information loss due to the low-resolution representation of the material domain. If the predicted path is discontinuous, we describe it as ``Incorrect'' regardless of the $F_1$ score. We show three examples of this behavior in Fig. \ref{fig:crackres}a. Our trained model achieves a mean $F_1$ score of $0.893$ on the continuous path group, a $0.693$ on the discontinuous path group, and a $0.870$ overall. Qualitatively, we consider $78.77\%$ of the predictions as ``Correct,'' $9.84\%$ as ``Plausible Alternative Paths,'' and $11.39\%$ as ``Incorrect.'' Table \ref{tab:crack-path} summarizes prediction results on the Mechanical MNIST Crack Path dataset.
    In contrast to excellent recent work on Physics-informed Neural Networks (PINNs) developed for fracture prediction tasks \citep{failure-machine-learning-2,failure-machine-learning-3}, our approach has the benefit of achieving accurate predictions on heterogeneous materials with varying material and micro-structural distributions. Looking forward, we anticipate that the evaluation framework presented in this Section could be used to compare our proposed MultiRes-WNet to related alternative approaches \citep{failure-machine-learning-1,one-d-crack-prediction2}. Specifically, we hope that our open source benchmark dataset combined with our well-defined quantitative and qualitative score metrics will enable direct comparison between our approach and the approaches developed by others.

    \section{Conclusion}
    \label{sec:conclusion}
    In this paper, our goal was to advance the state of the art in predicting mechanically driven full field quantities of interest in a supervised learning framework. To this end, we propose a MultiRes-WNet, a novel type of Convolutional Neural Network, for mechanics-based full field quantity of interest prediction. We first showcase the capabilities of the MultiRes-WNet in accurately predicting full-field displacements in simulations of heterogeneous hyperelastic material domains from the previously created Mechanical MNIST \cite{lejeune2020mechanical0} and Mechanical MNIST Fashion \cite{lejeune2020mechanical} datasets. 
    After this initial success, we realized that it was necessary to introduce a more challenging dataset to evaluate our proposed method. 
    Therefore, we created the Mechanical MNIST Crack Path dataset \cite{mech-mnsit-crack-path-openbu} through Finite Element simulations of a heterogeneous material domain undergoing quasi-static brittle fracture captured through the phase-field method.
    This significant extension to the Mechanical MNIST dataset collection \cite{lejeune2020mechanical0} allows us to study a different type of full-field QoI prediction -- full field crack path prediction. With this dataset, we show that by combining the MultiRes-WNet with an autoencoder, the MultiRes-Wnet is capable of making reasonably accurate crack path predictions on the Mechanical MNIST Crack Path dataset. Critically, we also introduce both an interpretable approach to reporting error on Mechanical MNIST Crack Path and an extended version of the dataset \cite{mech-mnsit-crack-path-dryad} that will allow other researchers to readily evaluate alternative metamodeling strategies against the baseline performance that we report here. 
    
    Despite the  remarkable performance of MultiRes-WNet in full-field displacement prediction, there are multiple potential directions for future work. For example, future research could explore methods to integrate boundary conditions as a model input \cite{graph-neural-nets}, or include information about the underlying physics of the problem \citep{pinn-main}. In addition, there is a significant need for methods that rely on smaller training set sizes \cite{lejeune2021exploring}. Furthermore, our results for the crack path prediction task can likely be improved upon. Critically, we anticipate that information loss in using low-resolution representations of the material distribution domain is a major contributor to prediction error. One promising avenue for future work would involve developing alternative methods that would function well with higher resolution input domains. In addition, future studies should explore alternative data curation and metamodeling strategies for crack path prediction that represent the crack path as a one dimensional entity \cite{one-d-crack-prediction1} rather than as a two dimensional damage field.
    In the extended version of the Mechanical MNIST Crack Path dataset \cite{mech-mnsit-crack-path-dryad} we provide $256\times256$ resolution information of displacements and damage fields for 20 steps throughout the simulation and full mesh resolution of displacements and damage fields at early stage of damage initiation and at the final displacement such that other researchers can explore these potential alternatives and make a direct comparison to the work presented in this paper. 
    We also briefly note that there are multiple additional supervised and unsupervised learning problems beyond the scope of this paper that could be formulated with these datasets.
    Although this work emphasizes the effectiveness of the MultiRes-WNet in the end-to-end mapping of a heterogeneous material to desired full-field QoIs, metamodeling in this area is still far from perfect. Therefore, we anticipate that by publishing both our dataset and our metamodel implementation under open-source licenses we will enable future research interested in mechanics-based metamodeling to exceed the baseline performance established here. 
    {\color{revs} And, perhaps even more critically, we hope that there will soon be many more open access mechanics based datasets available to further challenge the efficacy of our MultiRes-WNet and inspire new methodological approaches.}
    
    \section{Additional Information}
    \label{sec:acknow}
    The lite version of the Mechanical MNIST Crack Path Dataset used in this manuscript is available through the OpenBU Institutional Repository at \url{https://open.bu.edu/handle/2144/42757} \citep{mech-mnsit-crack-path-openbu}. The full version of the dataset is available through the Dryad Digital Repository \url{https://doi.org/10.5061/dryad.rv15dv486} \citep{mech-mnsit-crack-path-dryad}. The Mechanical MNIST and Mechanical MNIST Fashion Datasets are also available through the OpenBU Institutional Repository \url{https://open.bu.edu/handle/2144/39371}. All code used to generate and process Mechanical MNIST Crack Path is available on GitHub at \url{https://github.com/saeedmhz/Mechanical-MNIST-Crack-Path}. All code used to construct the MultiRes-WNet is also available on GitHub \url{https://github.com/saeedmhz/MultiRes-WNet}. 
    
    \section{Declaration of competing interest}
    The authors declare that they have no known competing financial interests or personal relationships that could have appeared to influence the work reported in this paper. 
    \section{Acknowledgements} 
    We would like to thank the staff of the Boston University Research Computing Services and the OpenBU Institutional Repository (in particular Eleni Castro) for their invaluable assistance with generating and disseminating Mechanical MNIST Crack Path. This work was made possible through start up funds from the Boston University Department of Mechanical Engineering, funds from the Division of Systems Engineering, the David R. Dalton Career Development Professorship, and the Hariri Institute Junior Faculty Fellowship. 
    \appendix
    
\begin{figure}[ht]
    \centering
    \includegraphics[width=\textwidth,height=\textheight,keepaspectratio]{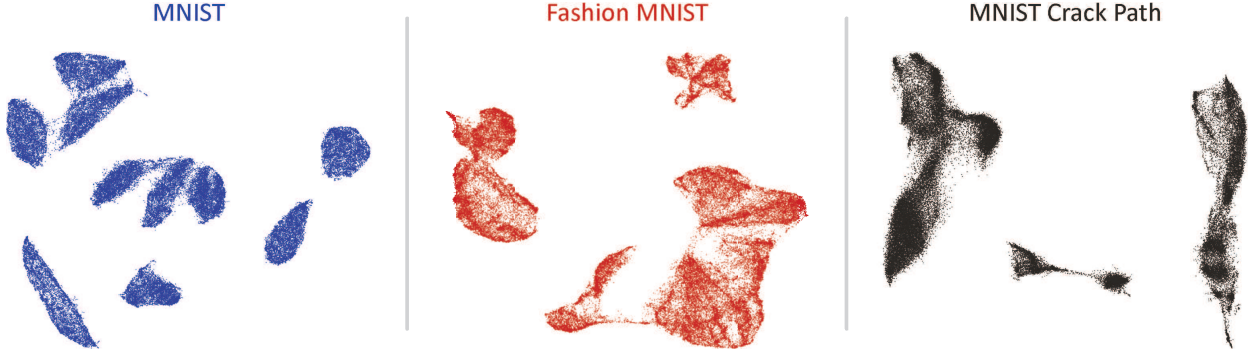}
    \caption{
    {\color{revs}
    Visualization of the input parameter space for the Mechanical MNIST dataset \citep{lejeune2020mechanical0}, the Mechanical MNIST Fashion dataset \citep{lejeune2020mechanical}, and the Mechanical MNIST Crack Path dataset \citep{mech-mnsit-crack-path-openbu} using the UMAP algorithm with default parameters {\fontfamily{qcr}\selectfont (n\_neighbors=15, min\_dist=0.1, n\_components=2, metric="euclidean")} \citep{mcinnes2020umap}.}
    }
    \label{fig:umap}
    \end{figure}

    {\color{revs}
    \section{Visualization of the Mechanical MNIST Collection Input Parameter Spaces}
    \label{apx:umap}
    In this Section, we provide supplementary information on the different input patterns in the Mechanical MNIST dataset collection. At present, there are three different input parameter distributions within the collection: (1) the MNIST handwritten digits \citep{mnist}; (2) the Fashion MNIST clothing images \citep{fashion-mnist}; and (3) the Mechanical MNIST Crack Path pattern which contains inclusions with center positions that are a random function of the Fashion MNIST bitmap pattern (see eqn. \ref{equ:rigidity-ratio}). To better visualize how these input parameter spaces are structured, we use the Uniform Manifold Approximation and Projection (UMAP) dimension reduction technique to project these high dimensional spaces into two dimensions \citep{mcinnes2020umap}. All three input parameter spaces are shown in Fig. \ref{fig:umap}. From this visualization, we can see that the MNIST, Fashion MNIST, and MNIST Crack Path datasets have distinctly organized input parameter spaces. Notably, the MNIST dataset is organized into $10$ quite distinct clusters corresponding to the $10$ classes in the original dataset \citep{mnist}, while the Fashion MNIST and MNIST Crack Path datasets are not, despite also being based on a dataset with $10$ classes \citep{fashion-mnist}. Looking forward, it will be important to further evaluate the predictive ability of the MultiRes-WNet for datasets with alternative material property distribution input spaces. For example, it would be broadly useful to the mechanics research community to develop multiple datasets that each address specific challenges in architected material design beyond the scope of the current Mechanical MNIST dataset.
    }
    \section{Details of Phase Field Model Implementation} 
    \label{apx:phase_field}
    In this Section, we summarize the process of data generation through Finite Element simulations of phase field fracture. In Section \ref{subsub:phase field}, we briefly introduce the phase-field method for quasi-static brittle fracture in general form. In Section \ref{subsub:chosen-methods}, we provide details of the specific form of the phase field method that we use in this work.
        \subsubsection{Phase Field Modeling of Quasi-static Brittle Fracture} \label{subsub:phase field}
\begin{figure}
        	\centering
        	\includegraphics[width=.75\textwidth,height=\textheight,keepaspectratio]{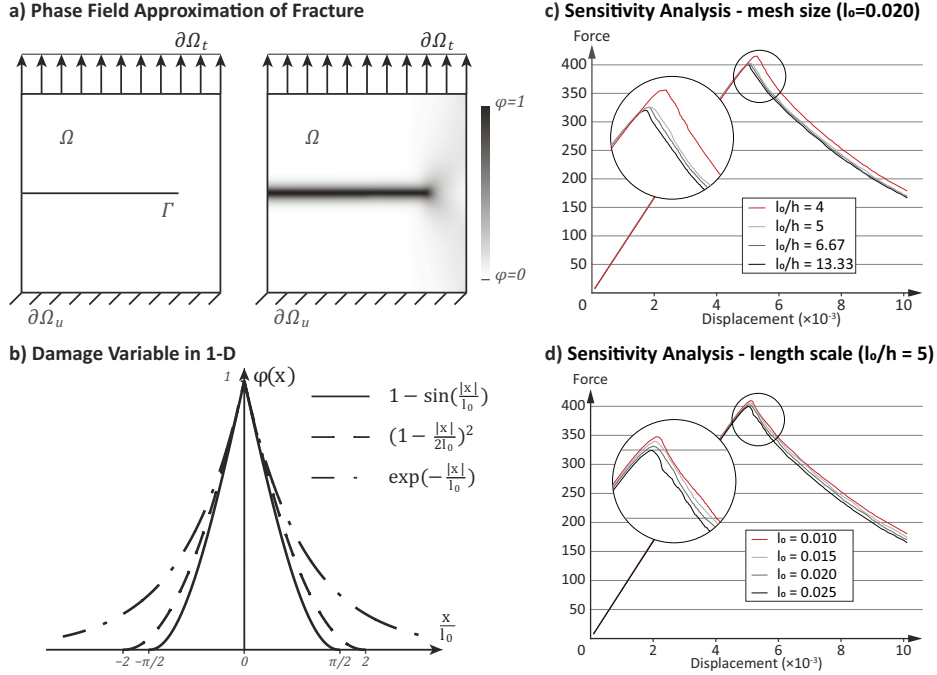}
        	\caption{Phase-field fracture modeling: a) shows a domain $\Omega \subset \mathbb{R}^d$ $(d=1,2,3)$ with a fracture occupying the region defined by $\Gamma \subset \mathbb{R}^{d-1}$ on the right in which $\partial\Omega_u$ and $\partial\Omega_t$ determine the Dirichlet and Neumann boundary conditions respectively and an approximation of the fracture with a continuous damage parameter $\phi$ where $\phi = 1$ represents the crack and $\phi = 0$ denotes the undamaged region on the left; b) one dimensional plots of three common choices for $\phi(x)$; c) force vs displacement curve for the problem domain of the Mechanical MNIST Crack Path Fig. \ref{fig:dataset}c for a fixed $l_0=0.020$ and different $\frac{l_0}{h}$ ($l_0$ is the length scale parameter and $h$ is the minimum mesh size); $\frac{l_0}{h} \geq 5$ produces reliable simulation results; d) force vs displacement curve for the problem domain of the Mechanical MNIST Crack Path Fig. \ref{fig:dataset}c for a fixed $\frac{l_0}{h} = 5$ and different length scale parameters. Note that for these illustrative plots, we simulate a homogeneous domain. In the Mechanical MNIST Crack Path dataset, $l_0=0.015$ and $h=0.003$ are chosen for generating the dataset.}
        	\label{fig:pf-intro}
        \end{figure}

        The core idea of the phase-field method in fracture modeling \citep{phase-field-review} is to approximate the discontinuity caused by fracture with a damage field variable. In Fig. \ref{fig:pf-intro}a, we show a domain $\Omega \subset \mathbb{R}^n$ $(n=1,2,3)$ with a crack occupying the region defined by $\Gamma \subset \mathbb{R}^{n-1}$ in which $\partial\Omega_u$ and $\partial\Omega_t$ define the Dirichlet and Neumann boundary conditions respectively. In Fig. \ref{fig:pf-intro}a, we show an approximation of the crack via a continuous damage parameter $\phi$ where $\phi = 1$ represents the crack and $\phi = 0$ denotes the undamaged region.  In Fig. \ref{fig:pf-intro}b, we show a one-dimensional illustration of three common choices for $\phi(\textbf{x})$. Following the paradigm established by Miehe et al. \citep{phase-field-miehe2010}, Wu \citep{phase-field-unified} introduced a unified method to approximate the crack by a regularized functional $\Gamma_d(\phi)$, Equ. \ref{equ:crack-approx2}, for quasi-static brittle fracture using the crack surface density function $\gamma(\phi,\nabla\phi)$:
        \begin{equation} \label{equ:crack-approx2}
            \Gamma \approx \Gamma_d(\phi) := \int_\mathcal{B}\gamma(\phi,\nabla\phi) \; \mathrm{d}\textbf{x} = \int_\mathcal{B} \frac{1}{c_0}\left(\frac{1}{l_0}\alpha(\phi)+l_0|\nabla\phi|^2\right) \mathrm{d}\textbf{x} \quad with \quad c_0 = 4\int_0^1\sqrt{\alpha(\phi)}d\phi,
        \end{equation}
        where $\alpha(\phi)$ is the geometric crack function and $l_0$ is an internal length scale parameter, and $\mathcal{B}\subset\Omega$ is the localization band over which the crack is smeared.
        The weak form of the governing equations for quasi-static brittle fracture, where we must find $\mathbf{u}\in\mathcal{U}_u$ and $\phi\in\mathcal{U}_\phi$, can be written as follows:
        \begin{equation} \label{equ:weak-form}
            \left\{
                \begin{array}{lr} 
                    \int_\Omega \left(\boldsymbol{\sigma} : \nabla_{sym}\delta \mathbf{u}\right) \mathrm{d}\mathbf{x} = \delta \mathcal{P} & \forall \delta \mathbf{u} \in \mathcal{V}_u \\
                    &\\
                    \int_\mathcal{B} \left( -Y \delta\phi + G_f \delta_\phi\gamma \right) \mathrm{d}\mathbf{x} = 0 \quad & \forall \delta \phi \in \mathcal{V}_\phi
                \end{array}
            \right.
        \end{equation}
        where $\delta\mathcal{P}$ denotes the power of the external body and surface forces, $G_f$ is the fracture toughness, and the stress tensor $\boldsymbol{\sigma}$ and the crack driving force $Y$ are defined as follows:
        \begin{equation} \label{equ:sigma-drive-force}
            \left\{
            \begin{array}{l}
                \boldsymbol{\sigma} := \frac{\partial\psi}{\partial\boldsymbol{\epsilon}} = \omega(\phi)\frac{\partial\psi_0}{\partial\boldsymbol{\epsilon}}\\
                \\
                Y:=-\frac{\partial\psi}{\partial\phi}=-\omega'(\phi)\frac{\partial\psi}{\partial\omega}=-\omega'(\phi)\bar Y
            \end{array}
            \right.
            \quad \mathrm{with} \quad \psi_0(\boldsymbol{\epsilon}) = \frac{1}{2}\boldsymbol{\epsilon}:\mathbb{E}_0:\boldsymbol{\epsilon}
        \end{equation}
        where $\psi_0(\boldsymbol{\epsilon})$, $\psi(\boldsymbol{\epsilon},\phi)$, and $\omega(\phi)$ are the initial strain energy function, the free energy density function, and the energetic degradation function respectively, and $\mathbb{E}_0$ is the elasticity tensor.
        Also, test and trial spaces are defined as follows:
        \begin{equation} \label{equ:test-trial-spaces}
            \begin{array}{ll}
                \mathcal{U}_u := \Big\{\mathbf{u}\;|\;\mathbf{u}(\mathbf{x})=\mathbf{u}^*, \; \forall \mathbf{x}\in\partial\Omega_u \Big\}, & \mathcal{V}_u := \Big\{\delta \mathbf{u} \;|\; \delta \mathbf{u} = \mathbf{0}, \; \forall \mathbf{x} \in \partial \Omega_u\Big\}\\
                &\\
                \mathcal{U}_\phi := \Big\{\phi \;|\; \phi(\mathbf{x})\in[0,1], \; \dot\phi(\mathbf{x}) \geq 0, \; \forall \mathbf{x}\in\mathcal{B}\Big\}, & \mathcal{V}_\phi := \Big\{\delta\phi \;|\; \delta\phi(\mathbf{x})\geq 0, \; \forall \mathbf{x} \in \mathcal{B}\Big\}
            \end{array}
        \end{equation}
        Finally, we note that following \citep{phase-field-unified} we consider the following generic expressions for $\alpha(\phi)$ and $\omega(\phi)$:
        \begin{equation}
        \label{equ:alpha-omega-generic-form}
            \begin{array}{c}
                \alpha(\phi) = \xi\phi+(1-\xi)\phi^2\\
                \\
                \omega(\phi) = \frac{(1-\phi)^2}{(1-\phi)^2+a_1\phi(1+a_2\phi)},
            \end{array}
        \end{equation}
        where $\xi$, $a_1$, and $a_2$ are constants.
    
    \begin{figure}[ht]
        	\centering
        	\includegraphics[width=\textwidth,height=\textheight,keepaspectratio]{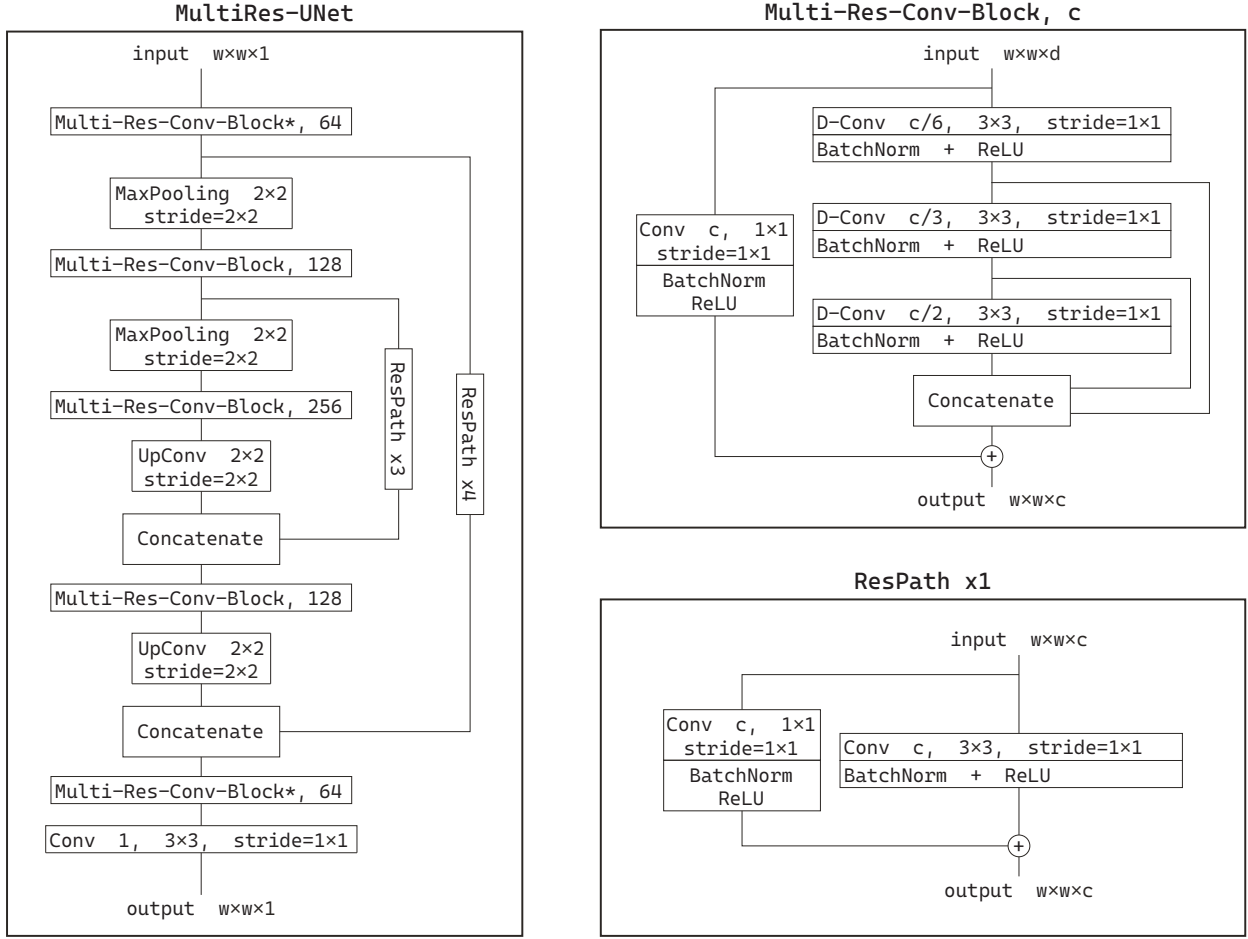}
        	\caption{The figure shows detailed architecture of MultiRes-UNet, Multi-Res-Conv-Block, and ResPath. Padding for each convolution is chosen such that the input and output arrays for that convolution have the same dimensions. D-Conv is used to refer to depthwise convolution. In Multi-Res-Conv-Block*, depthwise convolutions are replaced by regular convolutios.}
        	\label{fig:network-details}
    \end{figure} 

        \subsubsection{Details of the Phase Field Method Formulation} \label{subsub:chosen-methods}
        The model introduced in Section \ref{subsub:phase field} by the crack driving force defined in eqn. \ref{equ:sigma-drive-force} results in an isotropic phase-field damage model with identical tensile and compressive fracture behaviour. To address this issue, we follow the hybrid model described in \citep{ambati-hybrid,phase-field-unified}. As proposed in \citep{wu-length-scale-insensitive}, we replace the crack driving force formulation in eqn. \ref{equ:sigma-drive-force} by:
        \begin{equation} \label{equ:driving-force-alternate}
            Y:=-\frac{\partial\bar\psi}{\partial\phi}=-\omega'(\phi)\bar Y \quad \mathrm{with} \quad \bar \psi(\boldsymbol{\epsilon},\phi)=\omega(\phi)\bar\psi_0(\boldsymbol{\epsilon})=\omega(\phi)\left(\frac{1}{2E_0}\left(max\{0,\bar\sigma_1\}\right)^2\right)
        \end{equation}
        where $E_0$ is the Young's Modulus and $\bar\sigma_1$ is the major principal stress. Furthermore, to ensure crack irreversibility, we use the local history filed $\mathcal{H}(\mathbf{x},t)$ proposed in \citep{phase-field-miehe2010}:
        \begin{equation} \label{equ:history}
            \bar Y = \mathcal{H}(\mathbf{x},t):=\max_{t'}\bar\psi_0(\boldsymbol{\epsilon},t') \, . 
        \end{equation}
        Finally, we follow the optimal characteristic functions for length scale independent phase-field model proposed in \citep{wu-length-scale-insensitive} and choose the following values for parameters $\xi$, $a_1$ and $a_2$:
        \begin{equation} \label{equ:pf-main-params}
            \left\{\xi = 2, \; a_1 = \frac{4l_{ch}}{c_0l_0}, \; a_2=-\frac{1}{2}\right\} \quad \mathrm{with} \quad c_0=\pi \quad \mathrm{and} \quad l_{ch} = \frac{E_0G_f}{f_t^2}
        \end{equation}
        where $f_t$ is the failure strength. We note briefly that our full FEniCS \citep{alnaes2015fenics,logg2012automated} finite element implementation is available on GitHub, with a link provided in Section \ref{sec:acknow}. 
   
    \section{Details of our MultiRes-WNet implementation} 
    \label{apx:metamodel}
    To supplement the description of our MultiRes-WNet in Sections \ref{sec:metamodeling} and \ref{sec:results}, we have included additional detail on our implementation of the MultiRes-WNet network in Fig. \ref{fig:network-details}. Critically, note that the padding value for each convolution is such that the input and output arrays for that convolution have the same dimensions. In addition to this schematic illustration, please note that we have uploaded our code to GitHub under an open source license, see Section \ref{sec:acknow} for access information. 
    
    {\color{revs}
    \section{Comparing the MultiRes-WNet to Alternative Model Architectures}
    In this Section, we report a summary of our investigation comparing the performance of the MultiRes-WNet to alternative model architectures commonly used in the field. Essentially, we show the performance of other approaches that were not designed specifically for these mechanics-based datasets to make the relative benefits of the MultiRes-WNet clearer. In Section \ref{apx:comparison-disp-pred}, we compare the performance of four different neural network structures including MultiRes-WNet for displacement prediction. Then, in Section \ref{apx:comparison-1d-crack}, we summarize the results of an alternative approach for crack prediction based on a 1D representation of the crack path and a feed forward neural network architecture.
    
    \subsection{Full-field Displacement Prediction Comparison} \label{apx:comparison-disp-pred}
    For this comparison, we trained and tested all machine learning models on the Mechanical MNIST Fashion Equibiaxial Extension dataset, as it is the most challenging displacement-based dataset in our collection. The four Neural Networks trained for direct comparison are as follows:
    \begin{itemize}[noitemsep]
    \item FNN Feed-Forward Neural Network ($3$ convolutional, $2$ fully connected layers) with $5,832,240$ parameters
    \item UNet ($2$-level) with $1,863,489$ parameters
    \item WNet ($2$-level) with $3,726,978$ parameters
    \item MultiRes-WNet (see Section \ref{subsubsec:mrwn} for details) with $1,941,594$ parameters
    \end{itemize} 
    All four networks were trained using the Adam Optimizer for $200$ epochs with a learning rate of $10^{-2}$ for the first $100$ epochs and $10^{-3}$ for the last $100$ epochs. In contrast to the Mechanical MNIST Fashion results presented in Fig. \ref{fig:displacement-pred-histogram}d, we did not use data augmentation for the results reported in this Section. The results of the investigation are reported in Fig. \ref{fig:comparison-hists} and Fig. \ref{fig:comparison-errVSepoch}. As Fig. \ref{fig:comparison-errVSepoch} shows, for this dataset, the performance of the MultiRes-WNet is superior to all other alternatives considered for prediction in both the ``x'' and ``y'' directions. However, it is worth pointing out that the WNet, which has twice as many parameters as MultiRes-WNet, is the second best model and performs only slightly worse than MultiRes-WNet. Notably, a standard $UNet$ architecture has quite poor performance on these data, as shown in both the training/test error with respect to number of epochs curves in Fig. \ref{fig:comparison-errVSepoch}, and the histogram of test error with worst case sample prediction visualizations in Fig. \ref{fig:comparison-hists}. Summarized performance evaluation metrics are reported in Table \ref{tab:disp-pred-alternatives} for a clear quantitative comparison of the different models. All code used to construct these models and reproduce our results is available on GitHub \url{https://github.com/saeedmhz/MultiRes-WNet}.
    
     \begin{figure}[ht]
    \centering
    \includegraphics[width=\textwidth,height=\textheight,keepaspectratio]{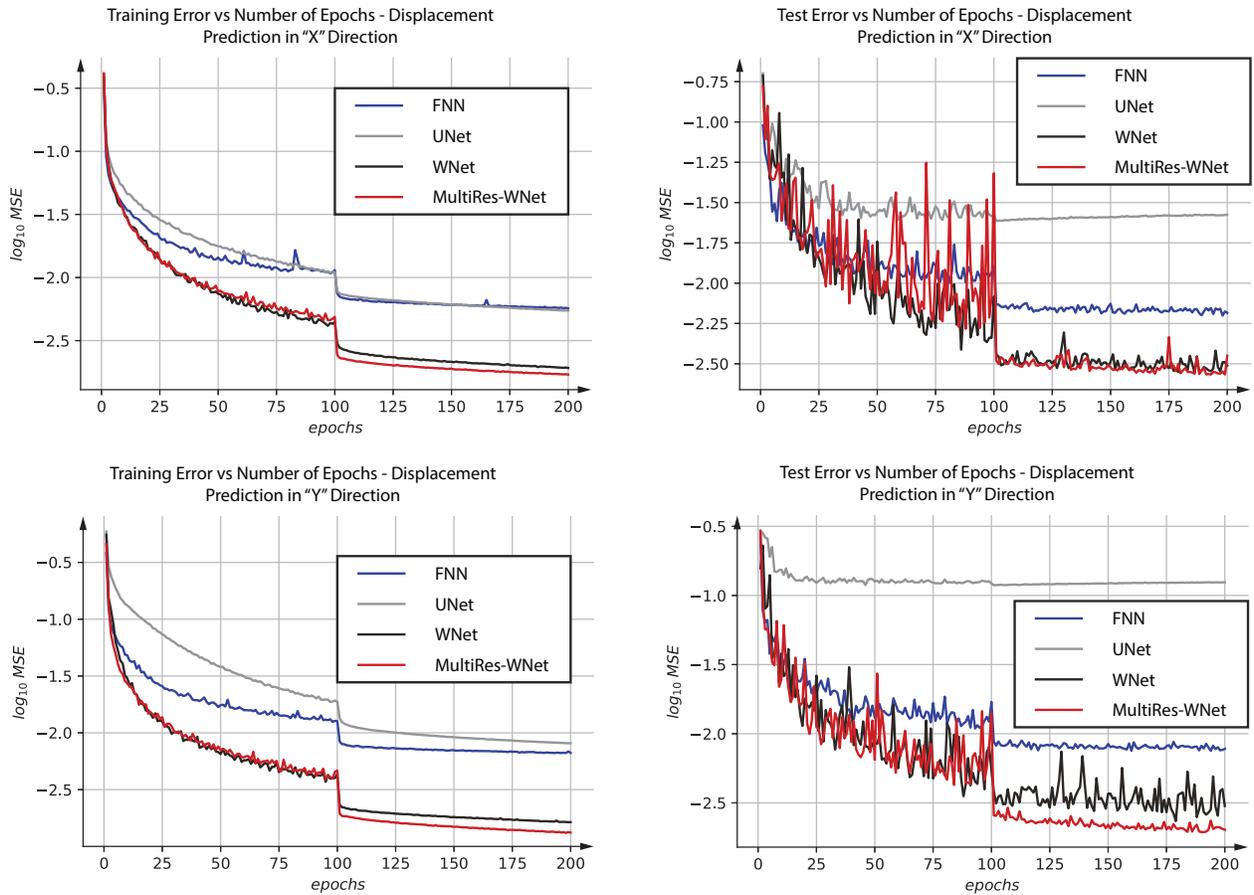}
    \caption{
    {\color{revs}
    Training (left) and test (right) error comparison for the FNN, UNet, WNet, and MultiRes-WNet models trained on the Mechanical MNIST Fashion - Equibiaxial Extension dataset for full-field displacement prediction in both the X (top) and Y (bottom) directions.}
    }
    \label{fig:comparison-errVSepoch}
    \end{figure}
    
    \begin{table}[ht]
      \centering
      \caption{
      {\color{revs}
      Performance comparison between the MultiRes-WNet and the three alternative Neural Networks investigated in this Section. All values reported correspond to test error. Please note that in contrast to the MultiRes-WNet results reported in Table \ref{tab:disp-pred-results}, these are results without data augmentation.}
      }
        \resizebox{0.8\textwidth}{!}{\begin{tabular}{ccccc}
        \toprule
              & \multicolumn{4}{c}{\textbf{Mechanical MNIST Fashion - Equibiaxial Extension}} \\
    \cmidrule{2-5}          & \textbf{MSE X Displacement} & \textbf{MSE Y Displacement} & \textbf{MAE} & \textbf{MAPE} \\
        \midrule
        \textbf{FNN} & 0.00629 & 0.00750 & 0.08657 & 2.027 \\
        \textbf{UNet} & 0.02438 & 0.11873 & 0.19251 & 4.507 \\
        \textbf{WNet} & 0.00279 & 0.00233 & 0.05035 & 1.179 \\
        \textbf{MultiRes-WNet} & 0.00270 & 0.00194 & 0.04867 & 1.139 \\
        \bottomrule
        \end{tabular}%
      \label{tab:disp-pred-alternatives}}%
    \end{table}%

    \begin{figure}[ht]
    \centering
    \includegraphics[width=\textwidth,height=\textheight,keepaspectratio]{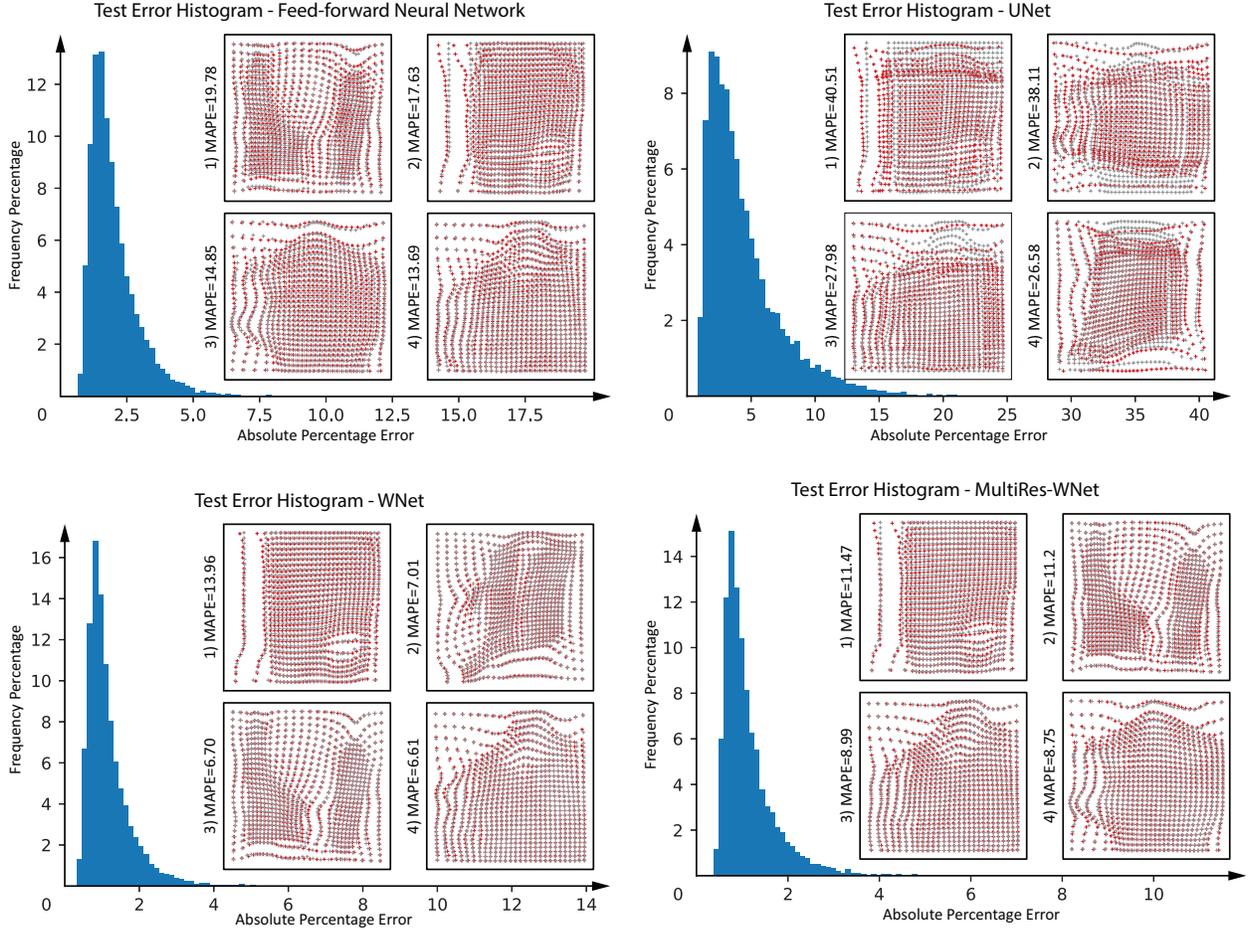}
    \caption{
    {\color{revs}
    Test error histograms for the FNN, UNet, WNet, and MultiRes-WNet models trained on the Mechanical MNIST Fashion - Equibiaxial Extension dataset. Each histogram is annotated with a visualization of the four worst predictions in the test set. Average test performance is summarized in Table \ref{tab:disp-pred-alternatives}.}
    }
    \label{fig:comparison-hists}
    \end{figure}

    \subsection{Crack Path Prediction Comparison} \label{apx:comparison-1d-crack}
    For the results reported in Section \ref{subsec:res1}, we used our MultiRes-WNet to predict the two dimensional damage field in the Mechanical MNIST Crack Path dataset \citep{mech-mnsit-crack-path-openbu}. The mean $F_1$ score with this approach was 0.87, and, qualitatively, we determined that 78.77 \% of predictions were ``Correct'' (defined as predicting a continuous path with an $F_1$ score greater than $0.85$). To give these results some additional context, we also explored an alternative crack path prediction approach that relies on a simpler method for crack representation (1D vector vs. 2D autoencoder) and a standard feed-forward neural architecture rather than the MultiRes-WNet. To achieve this, we first preprocessed the outputs of our Mechanical MNIST Crack Path dataset (2D damage fields) to obtain vectors that approximate 1D crack paths with the crack path $y$ coordinates sampled uniformly over the width of the domain. We note that for our Mechanical MNIST crack path dataset, the crack can be reasonably represented through this approach. However, implementing a similar naive approach would be nontrivial in more complex scenarios such as crack branching or multiple cracks. Then, we trained a feed-forward neural network (FNN) similar to the VGG-Net \citep{simonyan2015deep-vgg} (a set of CNNs for feature extraction followed by fully connected layers with $3,535,264$ parameters in total) on this processed dataset. We interpret the performance of this trained FNN via the two histograms illustrated in Fig. \ref{fig:comparison-1d-crack}. In Fig. \ref{fig:comparison-1d-crack}, the left histogram shows the $l_2$ distance between the predicted and true vectors representing the crack path in the test set. The right histogram is obtained by reconstructing the 2D damage fields from both the predicted and true 1D crack paths in the test set, and calculating the corresponding $F_1$ score (see eqn. \ref{equ:dice-score}). We briefly note that for 2D damage field reconstruction, we set the value of each pixel in the $256\times256$ domain to $1$ if the 1D crack curve crosses that pixel, and then offset the crack path obtained through this process by one pixel upward and downward in ``y'' direction while the rest of the pixels are zero. We chose to present the test error in two ways to better facilitate a fair comparison between the FNN vector prediction approach and our MultiRes-WNet damage field prediction. We also annotated each histograms with best, worst, and average predictions to aid in qualitative comparison. Comparing this alternative approach with our proposed method, we note that the predicted paths using this alternative approach are overly smooth, and even in the best cases predictions they fail to follow the complex irregularities that naturally exist in our crack path dataset. In addition, though it is difficult to perform a fair comparison between our proposed approach and this straightforward alternative, we note that our proposed approach achieves a much higher average $F_1$ score than the alternative approach ($0.87$ vs. $0.49$), and overall a much better predictive ability is seen in the representative examples in Fig. \ref{fig:crackres} and Fig. \ref{fig:comparison-1d-crack}. Since it would also be non-trivial to predict either crack bifurcation or multiple cracks using the vector approximation approach, we believe that our proposed approach will be more broadly applicable to others in the future. All code used to construct this alternative model and reproduce our results is available on GitHub \url{https://github.com/saeedmhz/MultiRes-WNet}.
    
    \begin{figure}[ht]
    \centering
    \includegraphics[width=\textwidth,height=\textheight,keepaspectratio]{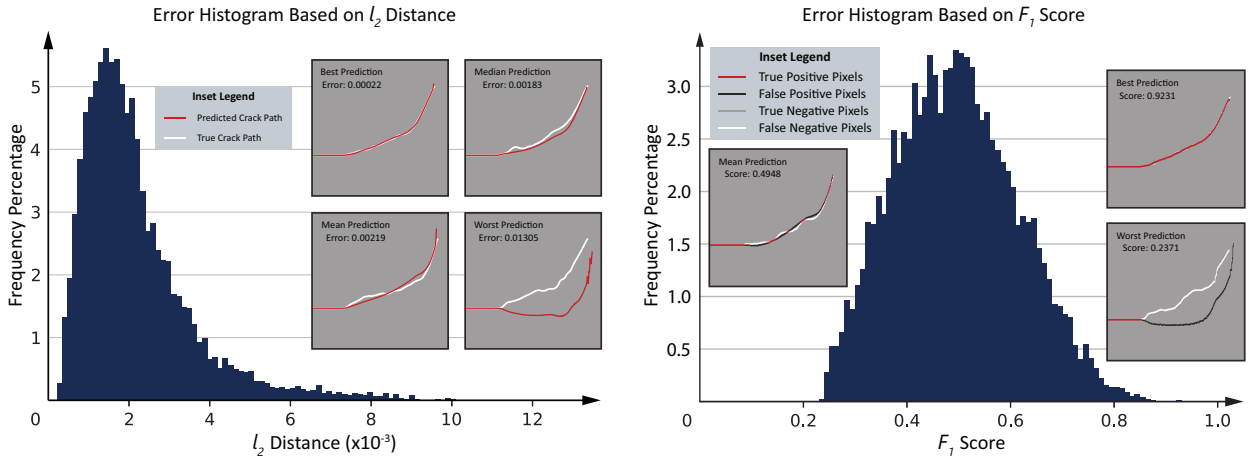}
    \caption{
    {\color{revs}
    Test error histograms for crack path prediction using the 1D vectorized representation of the crack path and a FNN model architecture. The left histogram measures the $l_2$ error based on true and predicted vector representations of the crack and the right histogram measures the $F_1$ error based on a reconstruction of the 2D damage field from both the true and predicted crack path vectors. The histograms are annotated with visualizations of the true and predicted crack paths for best, worst, and average predictions.}
    }
    \label{fig:comparison-1d-crack}
    \end{figure}
    }
    
    {\color{revs}
    \section{Predicting Full Field Strain with the MultiRes-WNet} \label{apx:strain-pred}
    The results presented in Section \ref{subsec:res1} show that the MultiRes-WNet is able to successfully predict displacement fields with high accuracy on the Mechanical MNIST \citep{lejeune2020mechanical0} and the Mechanical MNIST Fashion \citep{lejeune2020mechanical} datasets. To investigate the capability of the MultiRes-WNet in predicting strain fields in addition to displacement fields, we have extended our Mechanical MNIST Fashion dataset \citep{lejeune2020mechanical} to contain strain data. Specifically, these datasets now contain the deformation gradient tensors for each example sampled over the same uniform $28\times28$ grid as the displacement field. To demonstrate the efficacy of the MultiRes-WNet at predicting strain fields, we trained the MultiRes-WNet to predict the first component of the deformation gradient tensor ($\mathbf{F}_{11}$). For additional comparison, we also calculated $\mathbf{F}_{11}$ by differentiating both the true and predicted displacement fields sampled on the uniform $28\times28$ grid and compared the results of these three different versions of calculated $\mathbf{F}_{11}$ with the ground truth values of $\mathbf{F}_{11}$ obtained directly from the Finite Element simulations.
    Briefly, we used the {\fontfamily{qcr}\selectfont CubicSpline(x,y,axis)} function from the {\fontfamily{qcr}\selectfont scipy.interpolate} module to interpolate the displacement field from the given $28\times28$ displacement array sampled along the $x$ axis ({\fontfamily{qcr}\selectfont axis=0}) and used {\fontfamily{qcr}\selectfont CubicSpline.derivative(nu=1)} to compute the displacement gradient \citep{virtanen2020scipy}.
    We report the results of our investigations in Fig. \ref{fig:strain-pred}. The left histogram compares the test set error when differentiating the true displacement field to the test set error when differentiating the MultiRes-WNet predicted displacement field. The right histogram compares the test set error for direct prediction of $\mathbf{F}_{11}$ to the test set error when differentiating the MultiRes-WNet predicted displacement field. In all cases, the error is with respect to the ground truth computed directly from the Finite Element simulations.
    Our investigation shows that the test error for $\mathbf{F}_{11}$ prediction is significantly lower when we directly predict $\mathbf{F}_{11}$ compared to when we derive $\mathbf{F}_{11}$ from the predicted ($28\times28$) displacement field. However, the left histograms in Fig. \ref{fig:strain-pred} shows that this difference predominantly stems from differentiating an interpolation over a low-resolution displacement field rather than from error in the displacement predictions themselves. In brief, we found that direct prediction of $\mathbf{F}_{11}$ using the MultiRes-WNet architecture described in Section \ref{subsubsec:mrwn} without data augmentation was excellent, with $0.84$ mean percent error. And, we found that the prediction of $\mathbf{F}_{11}$ obtained from differentiating the MultiRes-WNet predicted displacement field had similar error to the prediction of $\mathbf{F}_{11}$ obtained from differentiating the true displacement field, with $3.93$ and $3.38$ mean percent error respectively.
    
    \begin{figure}[ht]
    \centering
    \includegraphics[width=\textwidth,height=\textheight,keepaspectratio]{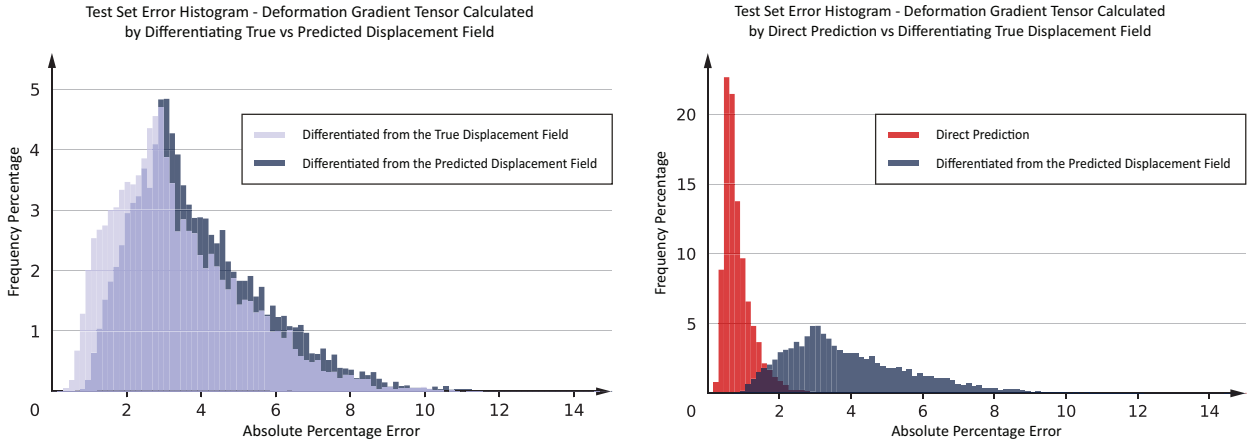}
    \caption{
    {\color{revs}Test error histograms for predicting $\mathbf{F}_{11}$ through both differentiating the displacement field and direct prediction. The left histogram compares the test set error when differentiating the true displacement field to the test set error when differentiating the MultiRes-WNet predicted displacement field. The right histogram compares the test set error for direct prediction of $\mathbf{F}_{11}$ to the test set error when differentiating the MultiRes-WNet predicted displacement field. In all cases, the error is with respect to the ground truth value of $\mathbf{F}_{11}$ computed directly from the Finite Element simulations.}
    }
    \label{fig:strain-pred}
    \end{figure}
    }
    
    {\color{revs}
    \section{Additional Information on the Definition of Qualitative Error Metrics for Crack Path Prediction} \label{apx:plausible-alt}
    In Section \ref{subsec:res2-crac}, in addition to using the $F_1$ score to quantitatively evaluate our model performance, we categorizes predictions into three qualitative groups: ``Correct''; ``Plausible Alternative Path''; and ``Incorrect.'' Illustrative examples are shown in Fig. \ref{fig:crackres}. To do this, we first labeled all predictions that lead to a discontinuous path as ``Incorrect'' predictions. Then, for the remaining continuous path predictions, we chose a cutoff threshold for the $F_1$ score where any prediction with a $F_1$ score less than the threshold was labeled as  a ``Plausible Alternative Path.'' The remainder of the predictions with both a continuous path and a $F_1$ score higher than the cutoff threshold were labeled as ``Correct'' predictions. In Fig. \ref{fig:plausible-alt}, we provide a plot that shows how sensitive our qualitative results are to the choice of the cutoff threshold. We selected $0.85$ for the threshold, and, as shown in Fig. \ref{fig:plausible-alt}, our selection of $0.85$ occurs just before a steep transition (this is also reflected in the histogram in Fig. \ref{fig:crackres}). Selection of a $0.85$ threshold was informed by both the curve shown in Fig. \ref{fig:plausible-alt}, and manual inspection of the data (see examples in Fig. \ref{fig:crackres}) which revealed that $0.85$ was a slightly conservative threshold for isolating crack paths that adhere to the ground truth and only have slight errors around their margin (see the zoomed in segments of Fig. \ref{fig:crackres}). With $0.85$ as the $F_1$ score cutoff we categorized $78.77$\% of the test set predictions as ``Correct.''   
    
    \begin{figure}[ht]
    \centering
    \includegraphics[width=0.475\textwidth,height=\textheight,keepaspectratio]{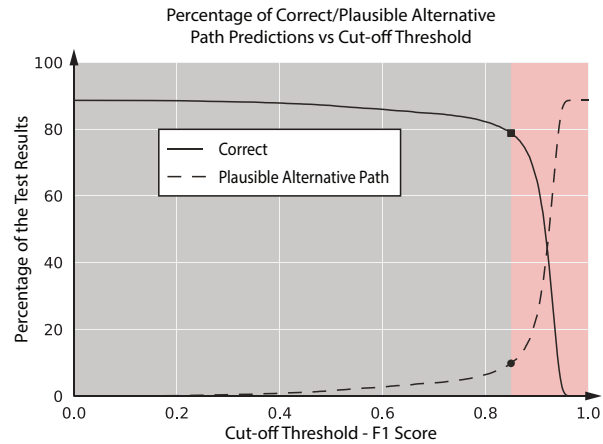}
    \caption{
    {\color{revs}Influence of the $F_1$ score cutoff in delineating between ``Plausible Alternative Path'' and ``Correct'' test predictions of our MultiRes-WNet on the Mechanical MNIST Crack Path dataset. In this work, we have selected $0.85$ as the minimum threshold for ``Correct'' prediction.}
    }
    \label{fig:plausible-alt}
    \end{figure}
    
    }

    \vspace{25mm}
    
    \bibliographystyle{plain}
    \bibliography{main}
   
\end{document}